# A Mathematical Approach to Constraining Neural Abstraction and the Mechanisms Needed to Scale to Higher-Order Cognition

Ananta Nair

## Abstract

Artificial intelligence has made great strides in the last decade but still falls short of the human brain, the best-known example of intelligence. Not much is known of the neural processes that allow the brain to make the leap to achieve so much from so little beyond its ability to create knowledge structures that can be flexibly and dynamically combined, recombined, and applied in new and novel ways. This paper proposes a mathematical approach using graph theory and spectral graph theory, to hypothesize how to constrain these neural clusters of information based on eigen-relationships. This same hypothesis is hierarchically applied to scale up from the smallest to the largest clusters of knowledge that eventually lead to model building and reasoning.

## Introduction

It can be agreed upon that the human brain is the best-known example of intelligence. However, not very much is known about the neural processes that allow the brain to make the leap to achieve so much from so little. The brain is an optimized system, capable of learning quickly and dynamically by creating knowledge structures that can be combined, recombined, and applied in new and novel ways. These knowledge structures, or schemas, help organize and interpret the world by efficiently breaking down and encoding information into numerous small blocks or abstract mental structures known as concepts (van Kesteren, Ruiter, Fernandez et al., 2012; Gilboa & Marlatte, 2017; van Kesteren & Meeter, 2020).

Concepts, when combined together in accordance with a relational memory system and inclusive of processes, such as maintenance, gating, reinforcement learning, memory, etc., form generalized structural knowledge (Whittington, Muller, Mark et al., 2019). A combination of these knowledge structures forms the building blocks that lead to the creation of world models for both model-based and model-free behavior (Chittaro & Ranon, 2004; Kurdi, Gershman, & Banaji, 2019). These well-organized structures, or models, further allow us to quickly process information, make decisions, and generalize to new domains from prior knowledge. New information can be encoded into the system to strengthen or change the schema, whereas post-encoding of this information can be consolidated, recombined, and retrieved, allowing for behaviors that fit with previous experiences or new behaviors that can be adapted to novel or changing situations (van Kesteren & Meeter, 2020).

Though the importance of schemas and mental abstraction has been recognized in the field of psychology and neuroscience, very little is known as to how the brain performs this difficult feat. This paper proposes a mathematical approach using graph theory and its subcomponent, spectral graph theory, to hypothesize how to constrain the neural clusters of concepts based on eigen-relationships. This same analysis is further applied to connections between clusters of concepts, the interaction of clusters that leads to structural knowledge and model building, and the interaction between models that results in reasoning. This paper will mainly focus on the functional connectivity of the brain, though a similar network of clusters can be implied at the structural level. Further, by drawing on past work, spectral graph theory will be discussed based on its relationship in determining functional connectivity from structural connectivity.

# Discussion

## Role of the brain's architecture in abstraction and its relation to higher-level cognition

It has long been believed that cognitive behaviors exist on a hierarchy (Botvinick, 2008; Badre & Nee, 2018; D'Mello, Gabrieli & Nee, 2020). According to Taylor, Burroni & Siegelmann (2015), the lowest levels of the pyramid structure represent inputs to the brain, such as somatosensory and muscular sensation, whereas the highest levels represent consciousness, imagination, reasoning, and other tangible behaviors, such as motor movements like tapping or painful stimuli. Though the progression of stimuli to thought has not been mapped, and limited evidence exits, it has been widely hypothesized that the brain organizes in global gradients of abstraction starting from sensory cortical inputs (Taylor, Burroni & Siegelmann, 2015; Mesulam, 1998; Jones & Powell, 1970).

This organization of information in global gradients and its continuous updating, storage, recombination, and recall can also be thought of as a continuous attractor model (Whittington, Muller, Mark et al., 2019). As the brain takes in information, these attractor states stabilize into common attractor states via error driven learning, with cleaned-up stable representations of the noisy input pattern (O'Reilly, Munakata, Frank et al., 2012). These stable representations employ self-organization learning to build knowledge structures and systems from which behavior can emerge. This allows for inputs at the bottom of the hierarchy to be scaled-up to the pinnacle layers of reasoning, consciousness, and other tangible behaviors of higher-level cognition.

## The interplay of structure and functional connectivity in the brain

Structural and functional connectivity are two important divisions that emerge to play different but interacting roles to achieve abstraction and subsequently higher-level cognition within the brain. Structural connectivity, much as the name suggests, is the connectivity that arises in the brain due to structure, such as white matter fiber connectivity between gray matter regions (Hagmann et al., 2008; Iturria-Medina et al., 2008; Gong et al., 2009; Abdelnour, Dayan, Devinsky et al., 2018). Contrarily, functional connectivity is concerned with the structure of relationships between brain regions, and typically does not rely upon assumptions about the underlying biology. It can be thought of as the undirected association and temporal correlations between two or more neurophysiological time series (such as those obtained from fMRI or EEG) (Chang & Glover, 2010; Abdelnour, Dayan, Devinsky et al., 2018).

The interaction between the brains structural and functional connectivity is of great interest in neuroscience but understanding the interplay has proven quite complex. Despite the existence of statistical models showing structural connectivity constrains functional connectivity (Honey et al., 2009; van den Heuvel et al., 2009; Abdelnour, Dayan, Devinsky et al., 2018), a full relationship between structure and function is not well developed.

## Mathematical approach to neuroscience

The brain is a complex system, and like other such architectures, necessitates the use of an interdisciplinary approach including mathematics, physics, and computational modeling. There currently exist many invasive techniques to understand the brain in animals. However, due to the sheer complexity and combinatorial dynamics of the brain, many of these techniques do not smoothly transition to human behavior. Due to these limitations, reliance must be placed on computational models and mathematical frameworks that can serve as systems for logical inference and hypothesis testing.

Mathematics especially provides a structure of systematic and logical thought that can be extended out to as many steps in the future as desired. Unlike computers, humans have limited quantitative abilities, and our logic generally does not extend past a finite number of steps. In addition, humans also find it difficult to consider and assimilate a large number of details across time-steps, something that can be easily addressed through equations. This feature is especially important when combining experiments to provide a theoretical framework. Unlike physics, which relies heavily on this method, neuroscience currently lacks sufficient theoretical frameworks to understand the brain and its emergent properties.

## An introduction to graph theories

Graph theory is a section of mathematics that studies graphs and other mathematical structures consisting of sets of objects, of which some may be related to each other. These graphs (G) as shown in Figure 1 and Equation 1, are made up of a set of edges (or lines, E) and vertices (or nodes, V). Graph theory studies the way in which these vertices are connected. Spectral graph theory is an extension of this theory and studies the properties of the Laplacian, edge, and adjacency matrices in relation to the graph.

$$G = (V, E), \text{ where } E \subset \{\{x; y\} \mid x, y \in V, x \neq y\} \qquad \text{(Eq. 1)}$$

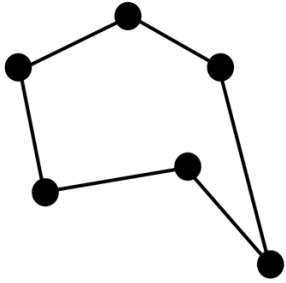

**Figure 1**: A graph "G". This graph consists of a set of vertices, or nodes 'V' that are connected by edges. 'E'. For simplicity, a d-regular graph is shown, where each node has the same number of connections or degrees 'd', where d = 2.

$$a_{ij} = \begin{cases} 1 \text{ if } \{i,j\} \in E \\ 0 \text{ otherwise} \end{cases} \quad \text{(Eq. 2)}$$

$$l_{ij} = \begin{cases} -1 \text{ if } \{i,j\} \in E \\ d(i) \text{ if } i=j \\ 0 \text{ otherwise} \end{cases} \quad \text{(Eq. 3)}$$

$$L = A - E \quad \text{(Eq. 4)}$$

$$A * x = \lambda * x \text{ where } x \in \mathbb{R}^v \quad \text{(Eq. 5)}$$

The adjacency matrix is a square matrix that represents which nodes are connected to other nodes. This can also be thought of as a manner of demonstrating connectivity. An example of a sample adjacency matrix is shown in Equation 2. The edge matrix is a square matrix that represents how many connections each node has. It is a binary matrix, with 1's representing a connected edge and 0 representing no connection between nodes. As expressed by Equation 4, subtracting the adjacency matrix from the edge matrix yields the Laplacian matrix. This matrix is a representation of a graph, and an example is shown in Equation 3. Similar to other matrices, these Laplacian matrices transform space and have eigenvectors and eigenvalues. In linear algebra, matrix transformations force vectors to scale and rotate. However there exist special vectors called eigenvector that are nonzero, real or complex, and stay unchanged or do not get knocked off their span regardless of the transformation (Equation 5). Eigenvalues correspond to the amounts by which eigenvectors are scaled.

In graph theory, eigenvalues and their corresponding eigenvectors are seen as solutions to optimization problems. Though eigenvalues do not possess enough information to determine graph structure, they are considered a good measure of numerical conductance, or connectivity. According to graph theory, this connectivity can be calculated using the second smallest eigenvalue. On the other hand, spectral graph theory uses a clustering technique to generate a number of clusters in large networks based on the Laplacian matrix and its eigenvectors and eigenvalues. These clusters allow us to divide the very large graphs studied in graph theory into smaller components to develop a better understanding of the problem space.

After computing the Laplacian matrix, a graph can be divided into two components by calculating a Fielder vector. This is the eigenvector corresponding to the second smallest eigenvalue of the Laplacian matrix of the graph and determines the Laplacian's algebraic connectivity. This vector has both positive and negative components, based on the graph's internal connections, and a sum of 0. These positive or negative values allow the graph to be divided into two distinct clusters based on its sign. Each of these clusters are well connected internally and sparsely connected externally. This is further explained in Equations 6 to 12 only for clarity. The process of dividing the graph into multiple clusters is addressed below.

$$if\ A\ is\ an\ operator : y = A * x\ and\ y_i = \sum_{j:(i,j)\ \in\ E} x(j) \qquad \text{(Eq. 6)}$$

$$A \text{ in quadratic form}: x^T = A * x = \sum_{(i,j) \in E} x(i) * x(j) \qquad \text{(Eq. 7)}$$

$$A * x = \lambda * x \ \& \ ||x|| = 1 \text{ then } x^T * A * x = \lambda \qquad \text{(Eq. 8)}$$

$$x^T L x = \sum_{(i,j) \in E} (x(i) - x(j))^2 \qquad \text{(Eq. 9)}$$

$$\lambda_1 = \min_{x \neq 0} \frac{x^T L x}{x^T x} \text{ and } v_1 = \text{argmin}_{x \neq 0} \frac{x^T L x}{x^T x} \text{ where } v_1 = 1 \qquad \text{(Eq. 10)}$$

$$\lambda_2 = \min_{x \perp v} \frac{x^T L x}{x^T x} \text{ and } v_2 = \text{argmin}_{x \perp v} \frac{x^T L x}{x^T x} \qquad \text{(Eq. 11)}$$

$$\lambda_k = \min_{S \text{ of dim } k} \max_{x \in S} \frac{x^T L x}{x^T x} \text{ and } v_k = \text{argmin}_{x \perp v_1, \ldots, v_{k-1}} \frac{x^T L x}{x^T x} \qquad \text{(Eq. 12)}$$

## **Mathematical approach to quantifying how structural brain connectivity leads to functional brain connectivity**

In Abdelnour, Dayan, Devinsky et al. (2018), the authors create a simple mathematical model based on graph theory to derive a relationship between structural connectivity measured using diffusion tensor imaging and functional connectivity measured from resting state fMRI. Although it is understood that a strong correlation exists between structural and functional connectivity and that functional connectivity is constrained by the structural component (Abdelnour, Dayan, Devinsky et al., 2018, Honey, Sporns, Cammoun, et al., 2009; Van Den Heuvel, Mandl, Kahn, et. Al, 2009; Hermundstad, Bassett, Brown, et al., 2013), there is no understanding of the relationship between the two types of connectivity.

Based on the theory that brain oscillations are a linear superposition of eignmodes (Raj, Cai, Xie, et al., 2020), Abdelnour, Dayan, Devinsky et al. (2018) demonstrate that structural connectivity and resting state functional connectivity are related through a Laplacian eigen structure. These eigen relationships arise naturally from the abstraction of time scaled and complex brain activity input into a simple novel linear model. In this model, a simple fitting procedure uses structural eigenvectors and eigenvalues to predict functional eigenvectors and eigenvalues. This is executed by using the eigen decomposition of the structural graph to predict the relationship between structural and functional connectivity. The linear spectral graph model only incorporated macroscopic structural connectivity without including local dynamics. As indicated in Mišić, Betzel, Nematzadeh, et al. (2015), though local dynamics are not linear or stationary, the emergent long-range behavior can be independent of detailed local dynamics.

This analysis is further used in a closed-form matter, where the eigenvectors are combined to build a model of full functional connectivity that is compared and verified against healthy functional and structural human data, as well as graph diffusion and nonlinear neural simulations. From their analysis, the authors are able to demonstrate that Laplacian eigenvectors can predict functional networks from independent component analysis at a group level. Though a strong relationship was seen at the level of the brain graph eigen-spectra, it was not seen at the node-pair level. However, this could be due to the model being completely analytical and using a single matrix exponentiation for whole brain functional connectivity, compared to using inter-regional couplings weighted by anatomical activity to influence neural node activity.

In summary, the paper captures the success of a graph model in confirming the linearity of brain signals, as well as drawing a relational connection between structure and function. The model further confirms, based on an analysis with human data, that long-range correlational structure exists in the brain through a mechanistic process based on structural connectivity pathways.

## Other work in network connectivity analysis

Although brain structure is imperative, functional connectivity is the star of the show. The brain is divided into regions based on specialized functionality, for example: neurons in the primary visual cortex or V1 detect edge orientation (Pellegrino, Vanzella, & Torre, 2004; Lee, Mumford, Romero, et. al, 1998). This manner of specialization for cross-modal communication and integration allows for the brain to perform abstract reasoning, reinforcement learning, memory maintenance, gating, and recall. All these processes individually and collectively contribute to higher level cognitive functioning, such as reasoning and conscious awareness, emerging from interactions across distributed functional networks.

In neuroscience, numerous analytical approaches have been used in an attempt to understand properties and make inferences about the brain. Graph theory, and subsequently the computational modeling in this domain, has focused on attempting to quantify specific features of the brain's network architecture. The work has modeled connectivity at the neuron level as well as examined simultaneous interactions among regions for fMRI experiments (Abdelnour, Dayan, Devinsky et al., 2018, Minati, Varotto, D'Incerti, et al., 2013). Although graph theory has been used to study the brain at a variety of levels, no work to date has thought to apply it to understanding how the brain forms concepts and recombines them to form generalized knowledge structures.

## The brain as an eigenproblem

Regardless of the task, the brain takes in some manner of environmental inputs (such as sound, light, somatosensory etc.). This information activates neurons in specific structural areas related to the task and those inputs further drive the activation of relevant functional areas. As found in Abdelnour, Dayan, Devinsky et al. (2018), structural and resting functional connectivity can be seen as a linear eigenproblem, with functional connectivity arising through a Laplacian eigen structure. Other research has also communicated the idea of understanding the brain as an eigenproblem (Hanson, Gagliardi, & Hanson, 2009). As with any complex system, it can be assumed that matrix computations and probabilities may not be the exact computation the brain uses but are tools that can be used to make sense of the mechanisms.

The following sections of this paper address the brain as an eigenproblem and use graph and spectral graph theory to make a connection to how the brain may deconstruct input into its smallest abstracted states. The problem will be addressed in the linear domain, to make ties to past work which has focused in this domain to maintain simplicity. A section has been included to addresses how this hypothesis translates to the non-linear domain.

**Constraining abstraction using graph theory across a single brain region**

As stated above, inputs to the brain activate certain groups of structural neurons, henceforth referred to as nodes, that in turn activate groups of functional nodes across a variety of brain regions. Across the brain this creates a network structure, where nodes from various regions of the brain connect with one another. Although there is a system level brain-wide network, it is simplest to initially think of activation in a particular brain region, for example, primary visual cortex or V1, as a network in itself. An example of this is shown in Figure 2 a.

Each node in a network has some probability of activating, based upon a threshold it must reach and some weighted connectivity with other nodes, as shown in Figure 2 b. When a person is initially unsure how to perform a task, such as identifying an object or performing a motor movement, these nodes will have some random probability of activating based on a random pattern of connection with other nodes. By applying graph theory to this random network, some algebraic connectivity can be calculated based on random eigenvectors and eigenvalues. Spectral analysis can further generate some random clusters for the graph. However, as the person learns the task and reduces this randomness through dopamine driven reinforcement learning and relational memory, some nodes develop stronger weights and connections whereas others develop weaker weights and connections. Through this learning, the network starts to develop established weights and connections, leading to a new and improved connectivity to emerge between nodes. This is discussed further below.

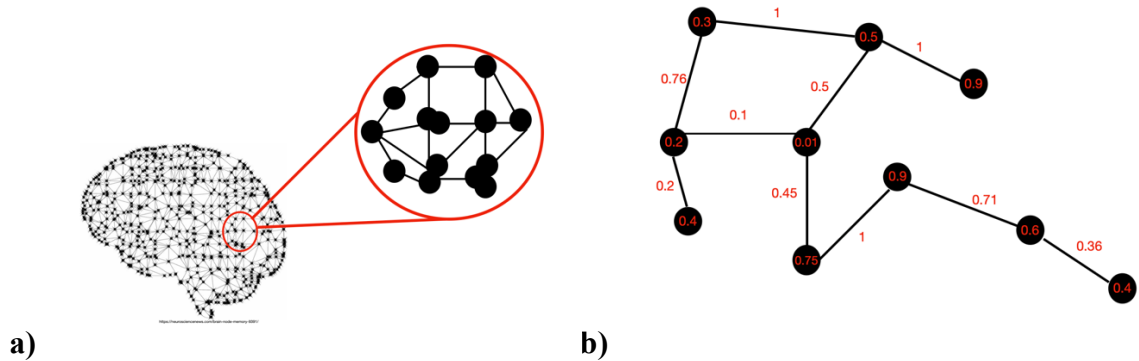

a)  b)

**Figure 2**: a) The left-hand side of the image showcases the entire brain as a network and the right-hand side shows how this large network can be simplified by focusing on a particular brain region, for example: V1. b) An example of nodes with activations (circles) and weighted connections (lines) in a network.

For this single brain region, graph theory can be used to calculate the algebraic connectivity of the network using the second smallest eigenvalue. Based on this calculated connectivity, an adjacency matrix, which demonstrates connectivity, and an edge matrix, which determines the number of connections per node can be calculated. Subtracting the adjacency matrix from the edge matrix yields the network's Laplacian matrix. Spectral graph theory, states that a clustering technique can be applied to this network to generate a number of clusters based on the Laplacian matrix and its eigenvectors and eigenvalues.

Though traditional methods divide the graph into two clusters, recursive bipartition or clustering using multiple eigenvalues and eigenvectors can be used to divide the graph into K-number of clusters. In recursive bipartition, a graph is first split into two clusters and those two networks are recursively split into smaller and smaller pieces. This can be done until arriving at the smallest informational pieces of the network. In clustering using multiple eigenvalues and eigenvectors, for every node in the graph a Laplacian matrix is calculated for which eigenvalue decomposition is undertaken, and the second, third, fourth and subsequently ascending smallest eigenvectors are calculated. This ensures every node of the graph is described by a small vector of values to which a clustering technique such as K-Means can be applied. However, instead of Euclidian distance, other techniques such as Manhattan or Fractional Distance will need to be utilized to account for high dimensional space (Aggarwal, Hinneburg, & Keim, 2001).

It has been often hypothesized that the brain uses some sort of clustering algorithm (Berry, Tkačik, 2020). The clustering technique used could be either of those mentioned above and may even vary depending on the brain areas or its structure/function. However, since brain regions depend on lateral and bidirectional dynamics, topological organization, and distributed representations, recursive bipartition seems a more likely technique to be employed by the brain from a computational time-cost standpoint. It also transfers more easily to the nonlinear domain, as discussed later. The clusters that arise from the spectral graph analysis, separate the once large interconnected network into a small group of nodes that are well connected internally and sparsely connected externally, as shown in figure 3 c. This method thus deconstructs the network of complex neural activation in a brain region into the smallest neuron activations, or neuron activations that represent the smallest pieces of information.

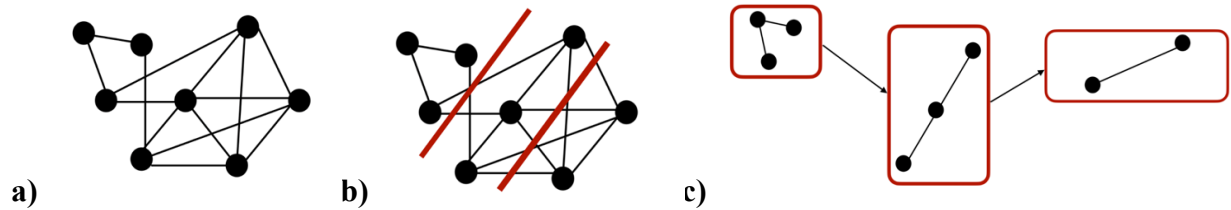

**Figure 3**: a) A whole network or graph network whose algebraic connectivity is determined by the second smallest eigenvalue of the graph. b) The division of a network using spectral graph theory into individual clusters representative of the smallest informational connections between nodes. c) A visualization of these clusters, that are well connected internally but sparsely connected externally.

## Constraining abstraction using graph theory across a brain-wide network

To reiterate so far, graph theory has been used to find network connectivity, and spectral clustering and graph theory have been used to take a network and subsequently divide it into its smallest components. By dividing this network located in a particular brain region out into its smallest clusters of neural activation, the hypothesis deduces that these clusters are the smallest neuronal structures of knowledge, defined above as concepts. For a single brain region, these clusters would be well connected internally and sparsely connected with other clusters in that same region. This would allow only certain clusters to become and remain active as well as be combined and recombined together, which is a dynamic possible due to the brain's large number of inhibitory neurons. This inhibition that allows only few neurons or clusters to activate at once, is especially important across brain regions for bidirectional connectivity, the formation of positive feedback loops, and sparse distributed representations.

To apply this analysis to the network of an entire brain, it can be inferred that small clusters of activation would emerge across all task-relevant regions. As an input enters the brain, say, an image of a chair for a characterization task, it would in parallel or in short succession activate a series of relevant brain regions; including those in the visual cortex, thalamus, prefrontal and parietal cortex, basal ganglia, and medial temporal lobe (Seger & Miller, 2010). For this large network, graph theory as indicated above could be utilized to determine connectivity amongst nodes, and spectral graph theory could then be further applied to determine clusters that reflect the smallest groups of neural activation or information. These neural clusters, representative of concepts, will have strong internal connectivity and sparse external connectivity within and across brain regions.

Across brain regions, distributed representations ensure multiple different ways of categorizing an input to be active at the same time. Successive levels of these representations are a main driving force for the emergence of intelligent behavior. Bidirectional connectivity uses these distributed representations to let clusters across various brain regions work together with other, albeit sparely connected, clusters to capture the complexity and subtlety needed to encode complex conceptual categories (O'Reilly, Munakata, Frank et al., 2012). Bidirectional connectivity is also essential to mechanisms like attention, and attributes to attractor dynamics, and multiple constrain satisfaction. This allow the network to stabilize into a stable and cleaned-up representation of a noisy input (O'Reilly, Munakata, Frank et al., 2012). In addition to these dynamics, error driven dopamine reinforcement learning and memory stabilize weights that develop between concepts across a brain-wide network. Based on weighting, concept clusters can be combined and recombined with other high weighted clusters to give rise to schemas. Creating schemas across various levels of dimensionality gives rise to generalized knowledge structures and models.

**<u>Levels of dimensionality of network cluster analysis</u>**

Computational neuroscience studies have shown that the brain employs sparse coding and dimensionality reduction as a ubiquitous coding strategy across brain regions and modalities.

Neurons are believed to encode high dimensional stimuli using sparse and distributed encodings, causing a reduction in dimensionality of complex-multimodal stimuli, and metabolically constraining the brain to represent the world (Beyeler, Rounds, Carlson, et. al, 2017).

The brain is charged with processing, storing, recalling, and representing high dimensional input stimuli to understand the world. To further add to this challenge, the brain is also constrained by metabolic or computational costs and anatomical bottlenecks. This forces the information stored in neuronal activity to be compressed into orders of magnitude smaller populations of downstream neurons (Beyeler, Rounds, Carlson, et al., 2019). A variety of dimensionality reduction technique have been hypothesized to do this, such as Independent or Principle Components Analysis (PCA), non-negative sparse PCA, non-negative semi joint PCA, non-negative sparse coding etc. (Zass & Shashua, 2006). Most of these methods use eigen-decomposition or other matrix-based decomposition techniques to reduce the number of variables required to represent a particular stimulus space.

In neuroscience, the number of variables implies the number of observed neurons. However, as these neurons are not independent of one another and span a variety of underlying networks, dimensionality reduction is used to find a small number of variables that can explain a network's activity. This is accomplished by determining how the firing rates of different neurons vary.

Manifolds are used to study varying degrees of dimensionality within the brain. A manifold is a topological space, that allows for real world inputs, such as images, sounds, neural activity, somatosensory sensation etc. to lie in low dimensional space while being embedded in high dimensional space. In neuroscience, it is believed that neural function is based on the activity of specific populations of neurons. These manifolds, obtained by applying dimensionality reduction techniques, serve as a surface that captures most of the variance of the neural data and represents neuron activity as a weighted combination of its time dependent activation (Luxem, 2019). For example, in Figure 4, the pink point can be thought of as a representation of sad and the green point can be thought of as a representation of happy in the manifold of facial expressions.

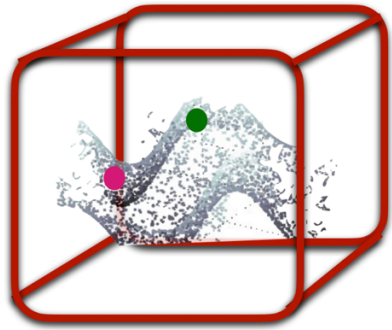

**Figure 4**: An example of a manifold. The manifold is the topological space in blue that lies in a low dimension while being embedded in a high dimensional space. This manifold is representative of human facial expressions, with the pink point being a representation of sad and the green point being representative of happy.

It is well accepted that the brain operates on a hierarchy, slowly adding components from increasing levels of dimensionality to arrive at higher level representations. As stated above, spectral graph theory breaks down a brain-wide network into its smallest pieces of neuronal information, or concepts, that exist at a reduced level of dimensionality. These concept clusters are combined and recombined with other clusters to yield schemas. This combination of clusters, or schemas, will exist at a higher level of dimensionality than its components and thus exist on a different manifold. Similarly, as schemas are combined and recombined across brain regions and levels of analysis, a hierarchy of generalized knowledge structures of varying degrees of complexity and dimensionality can be attained, such as plans, goals, outcomes, and eventually world models.

**Attractor dynamics and the formation of stable and learned clusters, schemas, and models**

From the previous sections, it has been discussed that graph theory can be used to calculate brain-wide network connectivity between nodes and spectral graph theory can be utilized to divide this network into its smallest components. A variety of mechanisms such as distributed representations, bidirectional connectivity, and dimensionality reduction, have also already been addressed as important dynamics that may push networks from a random state to form stable clusters, as well as attempt to scale representations to higher levels of cognition. However, the mechanisms that may account for the relational memory system that is hypothesized to be crucial in reorganizing concepts to form schemas and other higher-level structures of knowledge have not been discussed. Below is a list of mechanisms crucial to this system.

<u>Maintained activation</u> — sustained activity has been seen especially in brain areas that relate to high level cognitive processing, such as memory, attention, and control. Additionally, this continuous activation allows for training a variety of cortical areas that can exert their top down influences through bi-directional connectivity (O'Reilly, Munakata, Frank et al., 2012).

<u>Gating, Maintenance, and Working Memory</u> — the Basal Ganglia, including the Internal Global Pallidus, External Global Pallidus, Substantia Nigra, Subthalamic Nucleus, and Thalamus serves as a Go/No-Go system that determine what information is valuable (O'Reilly, Munakata, Frank et al., 2012). In working memory and task control, persistent neural firing is seen in the Basal Ganglia and its corresponding Basal Ganglia-Cortical loops. These loops are believed to guide neural processes as well as maintain and store representations in memory. The frontal cortex is responsible for robust active maintenance, while the Basal Ganglia contributes to selective and dynamic gating that allows frontal cortical memory representations to be rapidly updated in a task relevant manner (Frank, Loughry, & O'Reilly, 2001). In Working Memory tasks, the Basal Ganglia gates working memory representations into the Prefrontal Cortex to support executive functioning (Hazy, Frank, & O'Reilly, 2007).

Reinforcement Learning — this mechanism plays a critical role in the formation of network connections and the stabilization of network weights. It is well accepted that the Cortico-Basal Ganglia circuitry and Dorsal and Ventral Striata are critical for acquisition and extinction of behavior, as well as organized hierarchically in iterating loops (Yin, Ostlund, & Balleine, 2008). Dopamine, a key driver of reward behavior, functions as a global level mechanism for synaptic behavior modification (Glimcher, 2011) by comparing the predicted and actual value of reward received. Another important learning mechanism, XCAL, which addresses self-organizing and error driven learning, serves as a complementary dynamic (O'Reilly, Munakata, Frank et al., 2012).

Memory — The hippocampus, the center of memory, is known for being very sparsely connected, which allows it to preform pattern completion and one-shot learning (O'Reilly, Munakata, Frank et al., 2012). It is known to promote information encoding into long-term memory by binding, strengthening and reactivating distributed cortical connections, especially with the ventral medial prefrontal cortex (Van Kesteren, Fernández, Norris et al., 2010). Active maintenance and episodic memory are also helpful in organizing complex knowledge structures (Schraw, 2006; Moscovitch, Cabeza, Winocur et al., 2016).

**Applying graph theory in a non-linear domain**

As stated above, this paper addressed the brain as an eigenproblem in linear space to align with past work completed in the linear domain and maintain simplicity. This paper employs graph and spectral graph theory to mathematically represent how the brain's network deconstructs sensory inputs into small, abstracted states. Connections are also made to how these abstractions can be scaled up by the network to higher level cognition by addressing the mechanisms that are most likely key driving factors.

Though local dynamics in the brain are not by any means linear or stationary, emergent long-range behavior has been shown to be independent of detailed local dynamics (Mišić, Betzel, Nematzadeh, et al., 2015; Abdelnour, Dayan, Devinsky et al., 2018). However, this obviously does not apply to the level of neural dynamics, and it is believed this space operates in a nonlinear domain (Zhang, Li, Rasch et al., 2013; Amsalem, Eyal, Rogozinski et al., 2020). The same graph and spectral theory algorithms applied to a linear workspace can be expanded to the nonlinear domain. Ample research has and continues to be conducted in this field (Hein & Bühler, 2010; Bühler & Hein, 2009; Letellier, Sendiña-Nadal, & Aguirre, 2018).

To generalize graph theory to dynamic nonlinear networks, the dimensional space or number of variables of operation needs to be reduce for practicality. The matrix structures used in the linear domain, will be converted to Jacobian matrices. The Jacobian, in short, is a square matrix that is required for the conversion of variables from one coordinate system to another. This is accomplished, by creating a matrix for a set of equations that find areas where a nonlinear transformation looks linear. Using first order partial derivatives, the matrix sums up all the changes of the component vectors along each coordinate axis. These matrices are then generalized to the remaining nonlinear space. To calculate Jacobian matrices when using graph theory, a graph needs to be constructed in which the nodes are considered state variables and the links represent only linear dynamical interdependencies. This graph is used to identify the largest connected subgraph where every node can be reached from another node in the subgraph (Letellier, Sendiña-Nadal, & Aguirre, 2018). Generalizing standard spectral graph clustering to a nonlinear domain is much simpler and the graph p — Laplacian can be used to convert the space to a nonlinear eigenproblem.

The standard spectral clustering technique, that divides a network into two parts, can be used as a nonlinear operator on graphs. This would be valid for a standard graph p — Laplacian, where p = 2. However, for dividing the network into multiple clusters, p — spectral clustering can be utilized to consecutively split the graph until a desired number of clusters is reached. This can be reached through established generalized versions, such as ratio cut, normalized cut, or even multi-partition versions of the Cheeger cut (Bühler & Hein, 2009). Though the sequential splitting of clusters is the more traditional method, a nonlinear K-means clustering of eigenvectors and values can also be used in a non-linear space.

## Conclusions

The brain is an enormous, complicated system and till date not very much is known about the neural processes that divide complex real-world information into its smallest components. These smallest components or concepts are dynamically combined, recombined, and applied in new and novel ways and are believed to be integral in allowing the brain to swiftly and flexibly adapt to the world around it. This paper hypothesizes a mathematical methodology for quantifying these concepts. By visualizing the brain as a network graph with neurons as nodes, linked by weighted connectivity, graph theory can be used to determine its algebraic connectivity, and spectral graph theory can divide this connectivity into its smallest subcomponents with high internal connectivity and low external connectivity.

The smallest components that a network can be divided into give the smallest clusters of informational storage. These concept clusters exist at a low level of dimensionality and can be combined and recombined with a limited number of external clusters within and across brain regions based on sparse connectivity. This combination of clusters gives rise to schemas, which exist at a higher level of dimensionality. As schemas are further combined and recombined, with an increasing number of clusters across an increasing number of brain regions and dimensional representations, it is possible to attain a hierarchy of generalized knowledge structures with different degrees of complexity and dimensionality.

As with any complex unlearned system, the brain initially serves as a random network with random activations and connections. Over time, neurological mechanisms function as driving dynamics to allow for stability and weighted connectivity to emerge across brain regions. As shown in Figure 5, there exists a logical grouping to the brain's activity, which allows for high level cognition to emerge from abstract representations.

Structure, be it within or across regions, exists at the lowest level of the hierarchy and is a grouping of brain areas with similar specialized properties. This structure, as indicated above, is a precursor and serves as the source of functional activity. In the functional domain, abstractions are the smallest currency of information. The neurological clusters that form concepts have strong internal connectivity and weak external connectivity, which allows them to be connected, combined, and recombined only with a small number of other clusters within and across the brain. These combinatorics are driven by both functional connectivity and structural properties.

Although this paper focuses mostly on clustering in the functional domain, the same clustering can be applied to the structural network. The structural and functional networks would likely work in unison to achieve higher-level cognition.

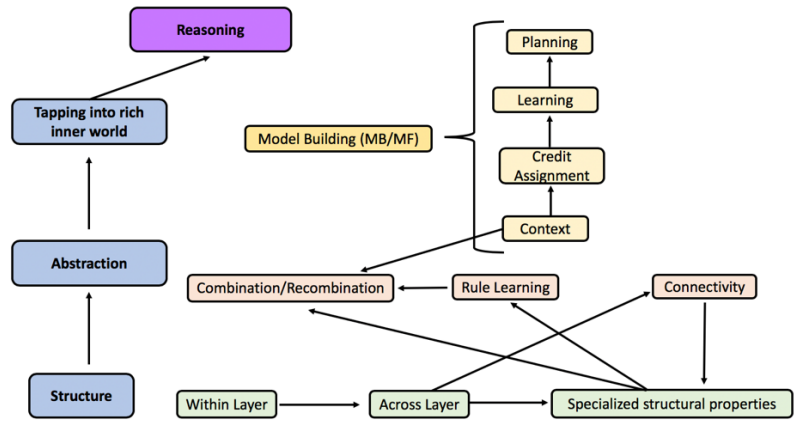

**Figure 5**: A diagrammatic grouping of brain activity into three groups. At the lowest level is structure, this activity within and across brain regions with specialized structure leads to functional activations. The smallest level of functionality exists in abstractions, and through structural, neural, and mechanistic processes allow for concepts to be combined and recombined to create higher order generalized structures of intelligence. At the third level, these higher-level generalized structures are combined with brain-wide high dimensional dynamics to form representations with an increasing degree of complexity and dimensionality. The highest-level representations from this level form models of the world that are critical components leading to human reasoning.

The rule learning components of these clusters are driven by neural factors such as bi-directional connectivity, distributed representations, attractor dynamics and multiple constraint satisfaction, and mechanistic components, such as maintained activation, gating and maintenance, reinforcement learning, and episodic memory as discussed above. These dynamics exist within and across brain regions, and determine how clusters can be combined and recombined. This allows for representations to gain an increasing level of dimensionality as an increasing number of clusters are combined.

Through rule learning, concepts are combined to form schemas and schemas are combined to form varying degrees of generalized knowledge structures, such as objects, categories, plans, actions, goals, outputs, and eventually models. The brain then hierarchically groups the highest dimensional representations of these knowledge structures into specific world models. When called, these model representations form the basis for conscious thoughts, ideas and predictions that are evaluated during reasoning.

## Future Work

To examine the viability of this hypothesis, future work is being conducted using computational modeling.